\pdfoutput=1
\documentclass[11pt]{article}
\usepackage[]{acl}

\usepackage{times}
\usepackage{latexsym}
\usepackage{color}
\usepackage{multirow}
\usepackage{forest}
\usepackage{tikz}
\usepackage{algorithm}
\usepackage{algpseudocode}
\usepackage{amsmath}
\usepackage{graphics}
\usepackage{subfigure}
\usepackage{mwe}
\usepackage{amssymb}
\usepackage{bm}
\usepackage{amsmath}
\usepackage{makecell}
\usepackage{tabularx}
\usepackage{url}
\usepackage{pgfplots}
\usepackage{CJKutf8}
\pgfplotsset{width=10cm,compat=1.9}
\usepackage{booktabs}
\usepackage{dsfont}
\usepackage{arydshln}
\usepackage{enumitem}
\usepackage[T1]{fontenc}
\usepackage[utf8]{inputenc}
\usepackage{microtype}
\usepackage[utf8]{inputenc} 
\usepackage[T1]{fontenc}    
\usepackage{url}            
\usepackage{booktabs}       
\usepackage{multirow}       
\usepackage{amsfonts}       
\usepackage{dsfont}     
\usepackage{nicefrac}       
\usepackage{microtype}      
\usepackage{tikz}       
\usepackage{tcolorbox}
\usetikzlibrary{shapes.misc}
\usetikzlibrary{shadows}

\DeclareRobustCommand{\tcgray}[1]{
\begin{tikzpicture}[baseline=(char.base)]
\node(char)[
  draw,fill=black!15,
  shape=rounded rectangle,
  text height=5pt,
  drop shadow={opacity=.5,shadow xshift=0pt,shadow yshift=-1pt},
]
  {\normalfont #1};
\end{tikzpicture}
}
\DeclareRobustCommand{\tcblack}[1]{
\begin{tikzpicture}[baseline=(char.base)]
\node(char)[
  draw,fill=black,
  shape=rounded rectangle,
  text height=5pt,
  drop shadow={opacity=.5,shadow xshift=0pt,shadow yshift=-1pt},
]
  {\color{white}{\normalfont #1}};
\end{tikzpicture}
}
\usepackage{bbold}

\title{On the Complementarity between \textit{Pre-Training} and \textit{Random-Initialization} \\for Resource-Rich Machine Translation}

\author{Changtong Zan$^{1}$\thanks{~~Work was done when Changtong was interning at JD Explore Academy.}~, \ Liang Ding$^{2}$\thanks{~~Liang Ding and Weifeng Liu are the corresponding authors.}, \ Li Shen$^{2}$, \ Yu Cao$^{3}$, \ Weifeng Liu$^{1\dagger}$, \ Dacheng Tao$^{2,3}$\\
$^{1}$College of Control Science and Engineering, China University of Petroleum (East China)\\ \ $^{2}$JD Explore Academy \ $^{3}$The University of Sydney\\
\texttt{b20050011@s.upc.edu.cn}, \texttt{dingliang1@jd.com}
}

\begin{document}
\maketitle
\begin{abstract}
Pre-Training (PT) of text representations has been successfully applied to low-resource Neural Machine Translation (NMT). However, it usually fails to achieve notable gains (sometimes, even worse) on resource-rich NMT on par with their Random-Initialization (RI) counterpart. 
We take the first step to investigate the complementarity between PT and RI in resource-rich scenarios via two probing analyses, and find that:
\tcblack{\small{1}} 
PT improves NOT the \textit{accuracy}, but the \textit{generalization} by achieving flatter loss landscapes than that of RI;
\tcblack{\small{2}}
PT improves NOT the confidence of lexical choice, but the \textit{negative diversity}\footnote{\citealt{li2020data} mentioned that the maximum likelihood estimation training treats all non-ground truth but semantic-relevant lexical predictions as being equally incorrect, potentially leading to the sharp lexical probability distribution.
Here we use this term to refer to the higher probabilities assigned to the negative tokens.} by assigning smoother lexical probability distributions than that of RI.
Based on these insights, we propose to combine their complementarities with a model fusion algorithm that utilizes optimal transport to align neurons between PT and RI. 
Experiments on two resource-rich translation benchmarks, WMT'17 English-Chinese (20M) and WMT'19 English-German (36M), show that PT and RI could be nicely complementary to each other, achieving substantial improvements considering both translation accuracy, generalization, and negative diversity. Probing tools and code are released at: \url{https://github.com/zanchangtong/PTvsRI}.
\end{abstract}

\section{Introduction}
Pre-training (\citealp[\textbf{PT};][]{Devlin:2018uk,liu2019roberta} has achieved tremendous success in natural language processing fields. Inspired by BERT~\cite{Devlin:2018uk}, recent works~\cite{song2019mass,bart2020,Liu:2020mbart} attempt to leverage sequence-to-sequence PT for neural machine translation (\citealp[NMT;][]{bahdanau2014neural,DBLP:journals/corr/GehringAGYD17,DBLP:journals/corr/VaswaniSPUJGKP17}) by leveraging a large amount of unlabeled (i.e. monolingual) sentences. 

While recent studies have empirically shown their benefit for the \textit{low-resource} translation task where the labeled (i.e. parallel) sentences are limited~\cite{bart2020,song2019mass,Liu:2020mbart}, we are generally confronted with \textit{resource-rich} scenarios, e.g. millions of parallel sentence pairs, in WMT evaluations~\cite{akhbardeh-EtAl:2021:WMT} and industries.
For these resource-rich tasks, PT becomes less effective (sometimes, even worse) than their Random-Initialization (\textbf{RI}) counterparts, for example, as \citealt{Zhu2020Incorporating,Liu:2020mbart} reported, the PT underperforms RI if improperly utilized or significant amount of bi-text data is given. 
However, there is limited understanding of: 
\tcgray{\small{1}} Why does PT fail compared to RI in terms of translation accuracy in high-resource settings?
\tcgray{\small{2}} What is the difference between the optimized PT and RI models?
\tcgray{\small{3}} How can we harmonize PT with RI? Can we just leverage their advantages?

\begin{figure*}[t] 
    \centering
    \includegraphics[width=1\textwidth]{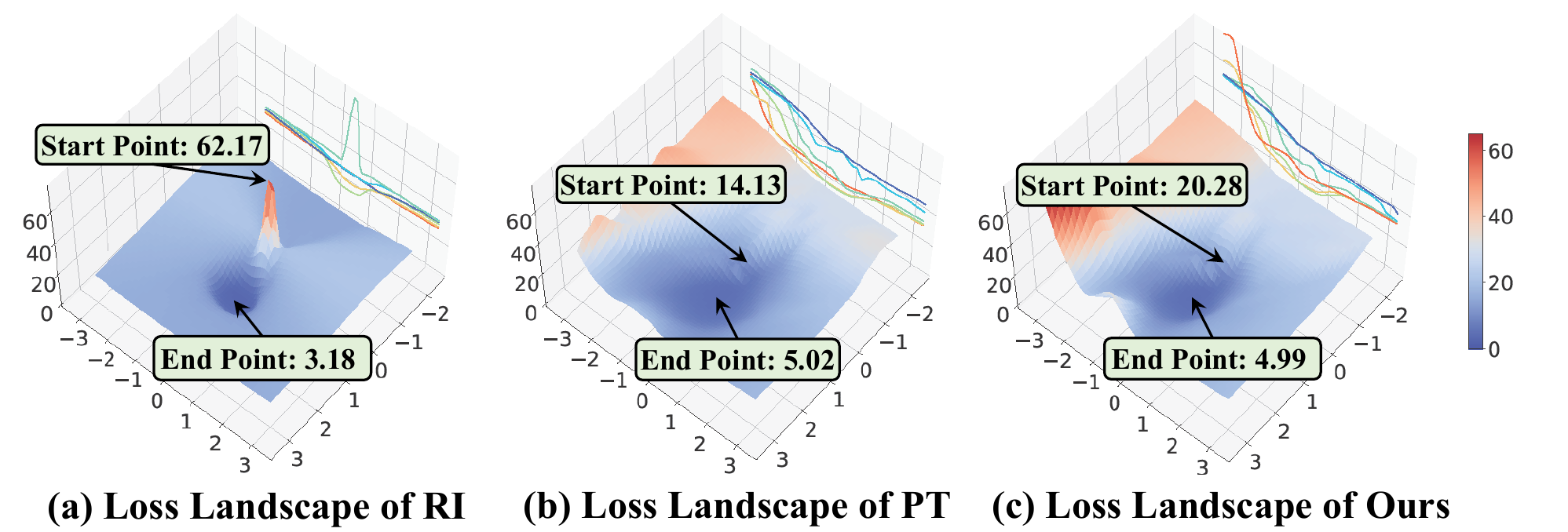}
    \caption{ Loss landscapes with contour lines on WMT19 En-De dataset. ``RI'' means the transformer training from scratch, while ``PT'' denotes the pretrained model mBART with finetuning. Both models have the same architecture.
    } 
    \label{fig:lls&ndi.}
\end{figure*} 

\begin{figure}[t] 
    \centering
    \includegraphics[width=0.51\textwidth]{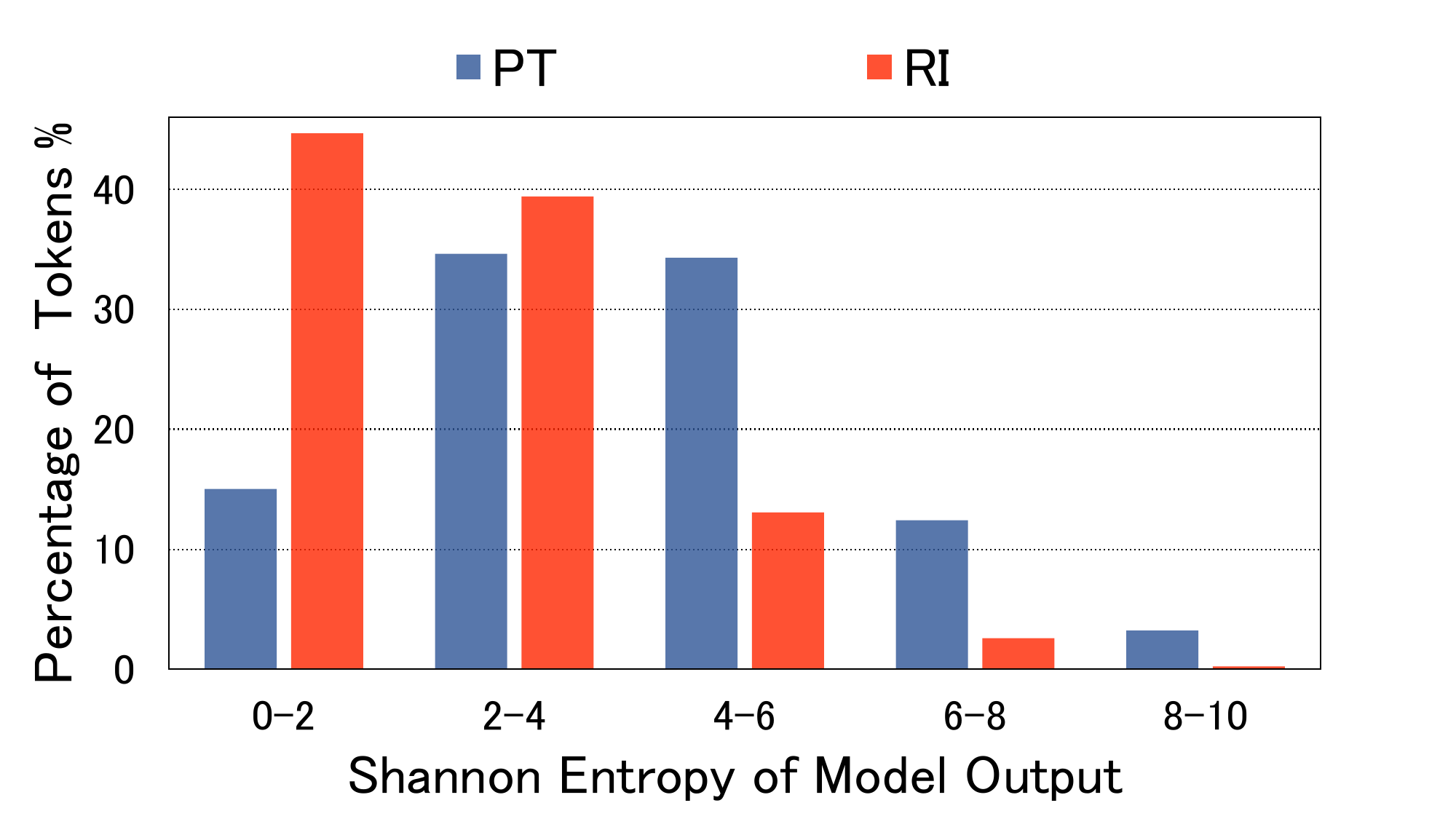}
    \caption{
    Token ratio distribution over difference entropy scales on WMT19 En-De test set. We compute the shanon entropy of the decoder output for each token.
    }
    \label{fig:ndi.}
\end{figure}

To this end, we introduce two probing analyses (i.e. loss landscape and lexical probability distribution) for the fully-optimized PT and RI models to investigate \tcgray{\small{1}} and \tcgray{\small{2}}, respectively. 
We find that: 
\tcblack{\small{1}} Supervised training with significant amount of parallel data, i.e. RI, is enough to optimize the model towards a better optimum point compared with PT, thus leading to better in-domain (same domain between test and training set) performance;
\tcblack{\small{2}} PT mainly contributes to the generalization ability because of its flatter loss landscape and smoother lexical probability distribution, thus resulting in better out-domain performance and diversified generation.
Motivated by this finding, we propose a simple combination method to reach \tcgray{\small{3}} by aligning the neurons and weights of PT and RI and then fusing them into a single model based on optimal transport~\cite{monge1781memoire,singh2020model}.
Experiments conducted on the WMT17 English-Chinese and WMT19 English-German benchmarks show that PT can nicely complement RI, leading to better translation accuracy, generalization, and negative diversity. 

\section{Experimental Setup} 
\label{sec:Setup}
\paragraph{Data} We conducted experiments on WMT17 English-Chinese (En-Zh) and WMT19 English-German (En-De) translation tasks, which are widely-used resource-rich benchmarks, and include 20M and 36M sentence pairs, respectively.

\paragraph{Setting} 
To make a fair comparison, all the model backbone architectures are the same as the pre-trained {\tt mBART.cc25}\footnote{\url{https://github.com/pytorch/fairseq/tree/master/examples/mbart}}. 
We tokenize with 250K SentencePiece model~\cite{kudo-richardson-2018-sentencepiece} of mBART and remove tokens that are not present in the downstream task from the vocabulary. For PT, we also initialize the model with trimmed embedding layers. 
For evaluation, we use SacreBLEU~\cite{post-2018-call} to measure the translation quality with default tokenizer.

\section{Understanding PT and RI} 
In this section, we aim to better understand the similarities and differences between PT and RI by introducing two probing analyses. The analyzed PT and RI models are trained on WMT19 En-De benchmark. 
We first present the loss landscape visualization of both models in Section~\ref{sec:loss_landscape}, and then show their difference in lexical probability distribution in Section~\ref{sec:ndi}. 

\subsection{Impact on Loss Landscape} 
\label{sec:loss_landscape} 
We follow~\citet{hao-etal-2019-visualizing} to visualize the two-dimensional (2D) loss surface, providing some insights on why PT fails compared to RI for in-domain accuracy while resulting better out-domain performance. 
Let $\theta_0$ denotes the initialized parameters, i.e., mBART parameters or randomly initialized parameters. $\theta_1$ and $\theta_2$ are the parameters of models trained on target dataset and auxiliary dataset respectively. We plot the loss surface with function as follows:
\begin{equation}
f(\alpha, \beta) = \mathcal{L}(\bm{\theta}_0+\alpha \bm{\delta}_1+\beta \bm{\delta}_2),
\label{eq:2d:loss}
\end{equation}
where $\mathcal{L}$ is the loss function, $\alpha$ and $\beta$ are scalar values represent current coordinate, $\bm{\delta}_1 = \theta_1 - \theta_0$ is the the optimization direction on target dataset, and $\bm{\delta}_2$ = $\frac{\left\| \theta_1 - \theta_0 \right\|}{\left\| \theta_2 - \theta_0 \right\|}$ $\left( \theta_2 - \theta_0 \right)$ denotes the normalized optimization direction on auxiliary dataset, $\left\| \cdot \right\|$ denotes the Euclidean norm. In our setting, we select mBART finetuned on WMT17 En-Zh data as the second axis to plot the loss landscape on WMT19 En-De benchmark.

Figure~\ref{fig:lls&ndi.}(a) and Figure~\ref{fig:lls&ndi.}(b) present the loss landscape with contour line of model training with PT and RI, where the left-side and right-side axis are corresponding to $\alpha$ and $\beta$ in Equation~\ref{eq:2d:loss} respectively.
We also mark the start point and the end point of training path. The start point denotes the loss of model without training, while the end point indicates the converged model loss.
Firstly, we can see that PT obtains flatter loss landscape than RI with contour line covers a wider area, which denotes that initialized with pre-trained parameters leads model to be more robust. 
This supports us to understand the generalization of PT from the perspective of the flatter optima.
Secondly, the start point of PT is much smaller than RI (14.13 vs. 62.17), which demonstrates that the knowledge learned during pre-training stage is also suitable for the downstream task.
We also observe that RI achieves the lowest loss at the end point (3.18). This shows that the RI model is well optimized on the in-domain data, as the large amount sentence pairs provide enough supervised signal.

\subsection{Impact on Lexical Prob. Distribution}
\label{sec:ndi} 
Previous work~\cite{jiang2019improving} shows that the lexical probability distribution is correlated with the diversity of generation, i.e., over-confidence in certain tokens leads to lower diversity~\cite{Miao2021PreventTL}. This finding inspires us to measure the density of PT and RI outputs. We use Shannon entropy of decoder output according to test set to probe this attribute. The smaller entropy represents less assigned probability density on incorrect tokens.

As shown in Figure~\ref{fig:ndi.}, we present how much percentage of character is involved in different intervals. 
We firstly see that most tokens prefer a relatively smaller entropy, which denotes that both models are well optimized, and predict tokens with higher confidence.
Besides, PT achieves a smoother distribution than RI with most outputs achieving relative higher entropy.
This suggests that PT could generate more diversiform translations, which is consistent with the report from ~\citet{wang-etal-2022-understanding}.

\section{Harmonizing PT and RI}
Based on the above experiments, we prove that PT mainly contributes to the generalization ability and RI leads to better in-domain performance. To combine the advantage of both models, we fuse both models via optimal transport based on activations following~\citet{singh2020model}.

\subsection{Model Fusion}
\paragraph{Discrete Optimal Transport} Given two discrete measures $\sum_{i=1}^{n} \alpha_i \delta_{\alpha_i}$ and $\sum_{i=1}^{n} \beta_i \delta_{\beta_i}$, where $\delta_{\beta_i}$ denotes the Dirac measures on $\beta_i$, and $\sum_{i=1}^{n} \alpha_i = 1$ and $\sum_{i=1}^{m} \beta_i = 1$. 
Then, with the  cost matrix $\textbf{C} \in \mathbb{R}^{n \times m}$, which computed with predefined cost function, the discrete optimal transport formulates:
\begin{equation}
\label{eq:ot-discret-problem-c4}
\text{OT} := \min_{\textbf{T} \in {\Pi}(\boldsymbol{\alpha}, \boldsymbol{\beta})} \langle\textbf{T}, \textbf{C}\rangle_F
\end{equation}
where $\langle \cdot, \cdot \rangle_{F}$ is the Frobenius product, and 
$\Pi(\boldsymbol{\alpha}, \boldsymbol{\beta}) = \{ \textbf{T} \in \mathbb{R}^{n \times m}: \textbf{T}\mathbb{1}_m=\boldsymbol{\alpha} \ \text{and} \ \textbf{T}^\top\mathbb{1}_n=\boldsymbol{\beta}\}$ 
is the set of all admissible transport matrix between the two measures (${\mathbb{1}}_n$ is a vector of ones of size $n$). 

\paragraph{Model Fusion} 
For layers $\ell_{ri}$ and $\ell_{pt}$ of RI and PT models respectively, we define the activation based probability measures over neurons as $\mu^{\ell_{ri}}=\big(\alpha^{\ell_{ri}}, \delta^{\ell_{ri}}\big)$ and $\nu^{\ell_{pt}}=\big(\beta^{\ell_{pt}}, \delta^{\ell_{pt}}\big)$, where $ \delta^{\ell_{ri}}$, $ \delta^{\ell_{pt}}$ are activations for given samples. And, we set uniform distribution for $\alpha$ and $\beta$. 

In term of the alignment procedure, we first align the weights of incoming edges $\mathbf{W}^{\ell,\, \ell-1}_{ri}$ by constructing convex combination with the weights of previous layer transport matrix $\mathbf{T}^{\ell-1}$, normalized appropriately via the inverse of corresponding column marginals $\beta^{(\ell-1)}$. After getting the aligned $\mathbf{\widehat{W}}^{\ell,\, \ell-1}_{ri}$, we compute the optimal transport map $\mathbf{T}^{\ell}$ between $\mu^{(\ell_{ri})}$ and $\nu^{(\ell_{pt})}$, which is formulated as Equation~\ref{eq:ot-discret-problem-c4}. And now, we align $\ell_{ri}$ with respect to $\ell_{pt}$,
\begin{equation}\label{eq:align}
\widetilde{\textbf{W}}^{(\ell, \,\ell-1)}_{ri}   \leftarrow \text{diag} \bigg(\frac{1}{{\beta^{(\ell)}}}\bigg) {\textbf{T}^{(\ell)}}^{\top} \mathbf{\widehat{W}}^{(\ell,\, \ell-1)}_{ri}.
\end{equation}
We will refer $\widetilde{\textbf{W}}^{(\ell, \,\ell-1)}_{ri}$ to the aligned weights of RI model, which we can directly add with $\textbf{W}^{(\ell, \,\ell-1)}_{pt}$. We carry out this procedure over all layers sequentially. Then, we fine-tune the fused model on corresponding dataset to obtain the final model.

\subsection{Results} 
As detailed in Section~\ref{sec:Setup}, we conduct the experiments on two widely used benchmarks WMT17 En-Zh and WMT19 En-De. To obtain the fused model, we set weights of 0.9 for PT and 0.1 for RI during model fusion and then fine-tune on the corresponding dataset. 

As present in Figure~\ref{fig:lls&ndi.}(c), we present the loss landscape of fused model. The start point is the fused model before fine-tuning. We can see that our model inherits the smooth loss landscape from PT and leads to smaller loss value.

The in-domain translation results are shown in Table~\ref{tab:main}. We present the results of PT and RI, which has the same network architecture, for comparison with our model. 
As seen, RI performs better than PT for two benchmarks (39.3 vs. 39.7 for En-De, 32.3 vs. 32.4 for En-De). This empirically verifies the claim we make in Section~\ref{sec:loss_landscape}.
Besides, our fused model achieves a better performance than PT (39.3~vs.~39.7 for En-De, 32.3~vs.~32.4 for En-De), which denotes that model fusion could promote the performance of PT.
while still under-performs RI  (40.1 vs. 39.7 for En-De, 32.9 vs. 32.4 for En-De). This may draw from that our approach can be seen as a trade-off of PT and RI.

\begin{table}[t]
\centering
\setlength{\tabcolsep}{5pt}
\scalebox{1}{
\begin{tabular}{lcc}
\toprule
\bf Model & \bf En-De & \bf En-Zh  \\ \midrule
\bf RI & 40.1 & 32.9 \\ \hdashline
\bf PT  & 39.3 & 32.3 \\
\bf ~~+Fusion & ~~~~~~~~$\text{39.7}^{\Uparrow +0.4}$ & ~~~~~~~~$\text{32.4}^{\Uparrow +0.1}$ \\
\bottomrule
\end{tabular}}
\caption{In-Domain translation quality of model trained on the WMT19 En-De and WMT17 En-Zh benchmarks. We evaluate models on official test sets. PT denotes the fine-tuned mBART, while RI indicts model trained from scratch.}
\label{tab:main}
\end{table}

\begin{table}[t]
\centering
\setlength{\tabcolsep}{5pt}
\scalebox{1}{
\begin{tabular}{lcccc}
\toprule
 & \bf PT & \bf RI & \bf Fusion & $\Delta$ \\\midrule
\bf 1) OD & 34.9 & 35.0 & 35.2 & \bf +0.2 \\
\bf 2) DuA & -12.1 & -16.7~~ & -12.3~~ & \bf  +4.4 \\
\bf 3) Multi-Ref & 75.4 & 75.8 & 76.0 & \bf  +0.2 \\
\bf 4) TTR & 19.1 & 19.1 & 19.2 & \bf+0.1 \\
\bottomrule
\end{tabular}}
\caption{Results of analysis experiments. ``OD'' (Out-Domain) evaluates the performance on the out-domain test set. ``DuA'' (Drop under Attack) reports the drop of BLEU when adding noise into source sentences. ``Multi-Ref'' is the BLEU score with multiple references. ``TTR'' denotes the Type-Token-Ratio. ``$\Delta$'' indicts difference between RI and Fusion.
}
\label{tab:analysis}
\end{table}

\subsection{Analysis}
We conducted several analyses to understand whether it inherits the advantages of PT. We evaluate our approach on four tasks, compared with PT model and RI model, including model generalization, performance drop under attack, translation diversity and type-token-ratio. Note that we can measure more properties, e.g. copy errors~\cite{liu-etal-2021-copying} and encoded linguistic knowledge~\cite{Hao2019ModelingRF,ding2020context}, but our aim is simply to show that PT and RI can be harmonized. All results are reported on the WMT19 En-De benchmark. For simplification, we list scores in one table.

\paragraph{1) Effects of Model Generalization}
To understand how our model performs on the out-domain data. We evaluate trained model on two different domains, including It and Medical, and report Out-Domain Score by averaging SacreBLEU scores. As it presents in the first line of Table~\ref{tab:analysis}, the fused model achieves $+0.2$ score compared with RI. This verifies the generalization ability of our model. 

\paragraph{2) Performance Drop under Attack}
To further demonstrate that our fused model could be more robust than RI. We attack the model by adding noises into source sentence during the inference stage and report the drop in score.
Specifically, we replace 10\% words with a special token. The less drop mean model is more robust.
As seen in the third line of Table~\ref{tab:analysis}, both models suffer a huge drop with the perturbed input. However, our model still performs better than RI (-29.6 vs. -31.2). This result shows that our model is less sensitive to noises, which leads to more robustness.

\paragraph{3) Effects of Translation Diversity}
Following \citet{wang-etal-2022-understanding}, we evaluate the translation diversity of models with multi-reference SacreBLEU. We inference on the test set released by \citet{pmlr-v80-ott18a}, which consists of ten additional reference translations for 500 English sentences extracted from WMT14 En-De test set. 
As shown in the second line of Table~\ref{tab:analysis}, our model achieves a higher multi-reference SacreBLEU score than RI. This empirically proves that our fused model could generate a more diverse translation with different word orders.

\paragraph{4) Effects of TTR} 
We also probe diversity of the lexical with \textbf{TTR} = $\frac{num. \ of \ types}{num. \ of \ tokens}$ following \citet{wei-etal-2022-learning}. The higher score of \textbf{TTR} inflects that the model tends to generate translation with different word types. As it present in the last line of Table~\ref{tab:analysis}, our fusion model achieves +0.16 gains over RI. This shows that our model could not only generate translation with more word orders but also predict words of more types.

\section{Conclusion and Future Work}
This paper provides several insights for pre-training (PT) and random-initialization (RI) on resource-rich machine translation. We carefully investigate the impact of PT and RI on loss landscape and negative diversity and reveal that RI leads to a lower loss value of in-domain data, while PT is more beneficial to flatter loss and smoother lexical distribution. We also propose to combine them with model fusion via optimal transport. Experimental results on WMT17 En-Zh and WMT19 En-De benchmarks show that model fusion is a nice trade-off method to utilize the complementarity of PT and RI.

For future work, we would explore other effective fusion algorithms~\cite{liu2022deep,wang2022meta} to combine the merits of PT and RI. Also, it will be interesting to investigate whether harmonizing PT and RI could complement existing data augmentation methods for NMT, e.g. back translation~\cite{liu2021complementarity}, data rejuvenation~\cite{jiao2020data}, bidirectional distillation~\cite{ding-etal-2021-rejuvenating} and training~\cite{ding2021improving}.

\section*{Acknowledgements}
We are grateful to the anonymous reviewers and the area chair for their insightful comments and suggestions. Changtong Zan and Weifeng Liu were supported by the Major Scientific and Technological Projects of China National Petroleum Corporation (CNPC) under Grant ZD2019-183-008, and in part by the National Natural Science Foundation of China under Grant 61671480.

\bibliography{main}
\bibliographystyle{acl_natbib}

\end{document}